\documentclass[11pt]{article}
\usepackage[left=1in,right=1in,top=0.9in,bottom=0.9in]{geometry}
\usepackage[T1]{fontenc}
\usepackage[utf8]{inputenc}
\usepackage{lmodern}
\usepackage{amsmath,amssymb,mathtools}
\usepackage{graphicx,booktabs,tabularx}
\usepackage{float}
\usepackage{microtype}
\usepackage{enumitem}
\usepackage{fancyhdr}
\usepackage[hidelinks]{hyperref}
\usepackage{xurl}
\usepackage{caption}
\setlist{nosep,leftmargin=1.5em}
\providecommand{\square}{\rule{0.6em}{0.6em}}
\title{\sffamily\bfseries\LARGE No Judgment Without a Reason\\[0.45em]
\large Counterfactual Receipts for Versioned AI Evaluators}
\author{Ye Chen\\
\small Alibaba Group
\and Weining Zhang\thanks{Corresponding author.}\\
\small Cheung Kong Graduate School of Business}
\date{}
\begin{document}
\maketitle

\begin{abstract}

An evaluator can keep the right label while changing for the wrong reason. This matters when evaluator outputs gate agent actions, route cases to review, or re-enter training as feedback. Most evaluation protocols record only whether the final label is correct. They do not test whether a change in judgment traces to a change in the evidence, the governing rule, or the authority under which the rule applies.

We make that account executable. An evaluator state is represented by three typed sources: grounds (evidence), norms (decision rules), and authority (rule applicability). Replacing any subset of the old sources with their revised versions produces an eight-cell counterfactual table, the \emph{judgment cube}. A \emph{judgment receipt} is the complete family of inclusion-minimal source replacements that reproduce the revised verdict. It does the work of a prime-implicant sufficient explanation, but for a version transition rather than a single decision, and in a formally weaker sense. Executed counterfactuals certify a receipt; a language model can only predict one. Exact enumeration, an antichain bound, and a worst-case query lower bound state what certification costs when the evaluator is a black box.

ReasonBench implements the framework in organizational-policy and logical-rule worlds with exact receipts: 19,520 cases from 845 independently split source units plus 7,200 paired controls. In a five-seed study frozen by content hash before any result was opened, direct receipt prediction with a Qwen3-1.7B backbone reaches 98.41\% exact receipt accuracy on the locked test. Predicting the full cube reaches 96.99\%; the paired difference is $-1.42$ percentage points (95\% cluster-bootstrap interval $[-2.87, -0.30]$). The frozen claim that cube supervision would improve receipt prediction is rejected; a three-seed Qwen3-0.6B replication agrees ($-1.26$ points $[-3.08, -0.04]$).

The finding that matters more is what high accuracy conceals. Under a meaning-preserving permutation of source order, the same receipt is recovered in only 54.8\% of direct predictions and 49.2\% of cube predictions. Trained on zero- or one-source changes and tested on two- or three-source changes, direct prediction keeps 93.75\% revised-verdict accuracy yet recovers only 7.16\% of changed-case receipts. A post-hoc study retrains with randomly permuted source sections. It raises order consistency from 55.9\% to 96.6\% for direct prediction at no significant cost on the frozen rendering, improves an untouched alternate rendering, and leaves the untargeted inverse relation broken. The cube's locked deficit widens to \(-5.71\) points \([-9.75, -1.94]\), driven by the six organizational clauses.

These failures, not the horse race between targets, carry the design lesson. A counterfactual table is a sound audit object, but a more structured target is not automatically a more robust learner. Reason-bearing evaluation should separate \emph{prediction} from \emph{certification} and report consistency under meaning-preserving transformations next to ordinary accuracy. Small models can propose receipts; executed counterfactuals certify them.

\end{abstract}

\section{A judgment and its account}

AI evaluators now sit inside operational loops. A judge model decides whether an agent action is released, which transaction is escalated, which trajectory a person inspects, and which examples flow back into training as reward signal (Zheng et al., 2023; Ouyang et al., 2022). In these positions a label does more than measure: it hands out permission and attention. The reliability of such evaluators, and the difficulty of overseeing them at scale, is a live concern (Bowman et al., 2022).

Suppose an evaluator changes from \nolinkurl{deny} to \nolinkurl{approve}. Three different events can explain the transition: the accepted facts changed, the governing rule was amended, or the rule gained authority over the case. A final-label benchmark erases the distinction. So does a free-text explanation that describes the revised case without showing what made the revised judgment differ from the old one.

The trouble is not a shortage of rationales. Models produce fluent explanations that fail to track the computation that produced the answer (Adebayo et al., 2018; Jacovi and Goldberg, 2020; Turpin et al., 2023), and their self-generated counterfactuals are unreliable (Mayne et al., 2025; Dehghanighobadi et al., 2025). The narrower gap is that versioned evaluators lack a \emph{testable} account of change. When two versions disagree, which declared change suffices to reproduce the new verdict? If two changes each suffice, does the record keep both? If the label is right but the recorded reason is wrong, where does that error show up?

We work from one constraint:

\begin{quote}

\itshape \textbf{A changed judgment incurs a debt of reasons.}

\end{quote}

Here \emph{reason} names neither a mental state nor a persuasive paragraph but a typed, counterfactually executable replacement in the evaluator's declared sources. The debt is discharged when the account lists every minimal replacement sufficient to reproduce the revised verdict.

The paper's contributions are, in order of what we believe survives scrutiny:

\begin{enumerate}

\item \textbf{A formalism that separates prediction from certification.} Grounds, norms, and authority define a local reason state; their old/new hybrids form a judgment cube ($2^3$ factorial intervention table over source replacements); the complete minimal sufficient hybrids form a judgment receipt. A model-emitted receipt is a prediction. Only executing the counterfactuals through a trusted adjudicator certifies one. The distinction is built into the definitions.

\item \textbf{Paired consistency as an evaluation primitive.} Each control transformation carries a known output relation, a metamorphic relation in the software-testing sense (Chen et al., 1998; Segura et al., 2016). Violating the relation certifies that at least one prediction in the pair is wrong, which yields an error lower bound (Proposition 5) with no gold labels for the transformed items, a diagnostic that unpaired robustness accuracy does not provide.

\item \textbf{ReasonBench.} 45 audited organizational policy clauses and 800 independently generated logical worlds, each with executable old and revised states, all eight hybrids, and the exact receipt family. Source units are split before prompt rendering, and every paired control is exactly labeled by construction.

\item \textbf{A negative result frozen in advance, kept visible, and the failures behind it.} Full-cube supervision was frozen internally as the favored hypothesis and lost. In hindsight the loss is not shocking: eight jointly scored cells are a strictly harder target than the receipt itself, so ``denser supervision helps'' was optimistic. The findings worth keeping are the ones the failed comparison surfaced: near-ceiling test accuracy alongside roughly 50\% consistency under source reordering, and revised verdicts surviving composition shifts that receipts do not.

\end{enumerate}

Along the way we give the failure modes short names with standard readings: \emph{drift} and \emph{inertia} (false-positive and false-negative change detection), \emph{misattribution} (wrong receipt on a true change), and \emph{reason debt} (selective risk on the attribution task). The vocabulary is small, but evaluator governance currently lacks it.

\textbf{Relation to PolicyPatch.} \emph{Eval Is an Institution} studies how an explicit policy amendment is admitted: checked, scoped, logged, made reversible (Zhang, 2026). The present paper begins after versioned sources exist and asks which changes in grounds, norms, or authority minimally account for a changed judgment. Formal object, labels, benchmark, models, and claims are all separate; no learned artifact or label crosses over. The two papers share a research program, not a result.

The claim also stops where execution stops. A receipt shows that a declared change was sufficient inside an executable evaluator. It does not show that the rule was just, the evidence true, or the authority legitimate. An inspectable account is the start of accountability, and only the start.

\section{Evaluator states}

\subsection{Grounds, norms, and authority}

Let evaluator version \(t\) be

\[
I_t=(G_t,N_t,A_t).
\]

\(G_t\) holds the grounds accepted for a case: observations, retrieved records, typed evidence. \(N_t\) holds the norms mapping accepted grounds to a disposition: thresholds, prohibitions, obligations, priorities, exceptions. \(A_t\) records which norms have jurisdiction over which actors, regions, products, channels, or procedural stages. In software terms the three are the evaluator's evidence inputs, its decision rules, and the rules' applicability conditions.

For case \(x\), a deterministic adjudicator, meaning an executable reference evaluator, returns

\[
J(I_t,x)=D\!\left(G_t(x),N_t|_x,A_t|_x,x\right)\in\mathcal Y,
\]

where \(N_t|_x\) and \(A_t|_x\) are the norms and authority relations that can affect \(x\). The local reason state, the declared input state for this case, is

\[
\rho(I_t,x)=(G_t(x),N_t|_x,A_t|_x).
\]

\textbf{Proposition 1 (functional dependence).} If \(D\) is deterministic and \(\rho(I_s,x)=\rho(I_t,x)\), then \(J(I_s,x)=J(I_t,x)\).

\emph{Proof.} The two calls to \(D\) receive the same arguments. Determinism gives the same output. \(\square\)

Read contrapositively: a changed deterministic judgment requires at least one changed local source. The converse fails, usefully. A source can change without moving the verdict, and these no-op revisions make good invariance controls, because a model may treat the mere presence of an amendment as a cue to flip.

Proposition 1 is elementary by design; it states functional dependence on declared inputs and nothing deeper. Its role is to fix an accounting obligation. If the organization declares these to be the locally sufficient sources of judgment, an unexplained change outside them points to an incomplete interface or implementation. The work lies in finding the complete minimal accounts, certifying them, and measuring how learned systems violate them.

The three-part state is an interface, not an ontology. Evidence can be socially produced, rules decide which evidence counts, and authority may be contested. None of that goes away; it has to show up at an explicit interface before any account can cover it. A deployment may expose more or fewer source types, and the construction below generalizes to \(k\) types at cost exponential in \(k\).

\subsection{Authority as a jurisdiction gate}

Authority is not another rule. A norm says what follows \emph{if it governs}; authority says whether it may govern this case. Collapsing the two turns a lack of jurisdiction into a decision on the merits.

ReasonBench makes the distinction executable. Let \(\delta_n(g,x)\in\{\texttt{approve},\texttt{deny}\}\) be the merits disposition under norm \(n\), and let \(a(n,g,x)\in\{0,1\}\) indicate jurisdiction. Then

\[
D(g,n,a,x)=
\begin{cases}
\texttt{refer}, & a(n,g,x)=0,\\
\delta_n(g,x), & a(n,g,x)=1.
\end{cases}
\]

Here \texttt{refer} is an abstention: the case escalates instead of receiving a merits ruling. Norms act at the first order; authority is a second-order gate on whether the first-order mapping may speak. This is how the benchmark treats authority; we claim no wider legal theory.

The three interfaces separate questions that a single prompt tends to blur: what is accepted as the case, what ought to follow, and who or what makes that norm operative here. Each has its own version history and its own failure mode.

\begin{table}[!htbp]
\centering
\small
\setlength{\tabcolsep}{4pt}
\resizebox{\textwidth}{!}{%
\begin{tabular}{llll}
\toprule
Source & Question at the interface & Nearest CS reading & Example revision \\
\midrule
Grounds \(G\) & What happened, as accepted? & evidence; input record & A verified timestamp replaces an estimate \\
Norms \(N\) & What follows from the grounds? & decision rule & A refund window moves from 14 to 30 days \\
Authority \(A\) & May this norm govern this case? & rule scope; applicability & A regional office gains cross-border jurisdiction \\
\bottomrule
\end{tabular}%
}
\end{table}

\section{Judgment receipts}

\subsection{The judgment cube}

Take an old evaluator \(I^0\) and a revised evaluator \(I^1\). For each source type, choose the old or the revised value. With \(\mathcal K=\{G,N,A\}\), let \(I^S\) use revised components for \(S\subseteq\mathcal K\) and old components otherwise. The \emph{judgment cube} for case \(x\) is

\[
Q_x(S)=J(I^S,x),\qquad S\subseteq\mathcal K,
\]

a \(2^3\) factorial table over source replacements; every cell comes from executing that hybrid evaluator.

The definition needs \emph{interface compatibility}: every hybrid must be a well-typed input to the same adjudicator contract. ReasonBench guarantees this by construction. A production change that renames fields, changes units, or shifts the meaning of a source needs an explicit migration first; otherwise the cube is undefined.

The endpoints \(Q_x(\varnothing)\) and \(Q_x(\mathcal K)\) say whether the verdict changed. The six interior cells say which single or joint replacements reproduce the revised outcome, and they expose interactions that endpoints and one-at-a-time ablations miss.

\subsection{Complete minimal accounts}

When the endpoints differ, the \emph{judgment receipt} is

\[
\mathcal R_x=
\left\{
S\subseteq\mathcal K:
Q_x(S)=Q_x(\mathcal K)
\;\land\;
\forall T\subsetneq S,\;
Q_x(T)\ne Q_x(\mathcal K)
\right\}.
\]

In classifier-explanation terms, a receipt collects minimal reproducing intervention sets for a version-to-version transition. It is analogous to prime-implicant sufficient reasons (Darwiche and Hirth, 2020), but formally weaker: without monotonicity, a superset of a sufficient replacement need not reproduce the revised verdict, so receipt members are not implicants. The candidate interventions also differ: typed source replacements, not feature values in one decision. The receipt is a family, not a chosen member. If replacing the grounds alone or the norm alone each reproduces the revised verdict, then \(\mathcal R_x=\{\{G\},\{N\}\}\), and recording only one suppresses an equally minimal account.

Three derived quantities recur:

\begin{samepage}

\begin{itemize}

\item \emph{reason complexity}: \(\min_{S\in\mathcal R_x}|S|\), the size of the smallest sufficient replacement;

\item \emph{receipt multiplicity}: \(|\mathcal R_x|\);

\item a transition is \emph{receipt-ambiguous} when \(|\mathcal R_x|>1\).

\end{itemize}

\end{samepage}

Ambiguity here is multiplicity within the declared interface, without legal or metaphysical weight. If the endpoints agree there is no receipt to give, even though sources may have changed.

Minimal sufficiency itself is not new, and non-Boolean features are already handled in sufficient-reason theory (Ji and Darwiche, 2023). What the receipt changes is the object: the unit of explanation is a version transition, the candidate interventions are typed source replacements, the full minimal family is preserved, and authority is a separately executable source.

\begin{figure}[tbp]
\centering
\includegraphics[width=0.90\textwidth]{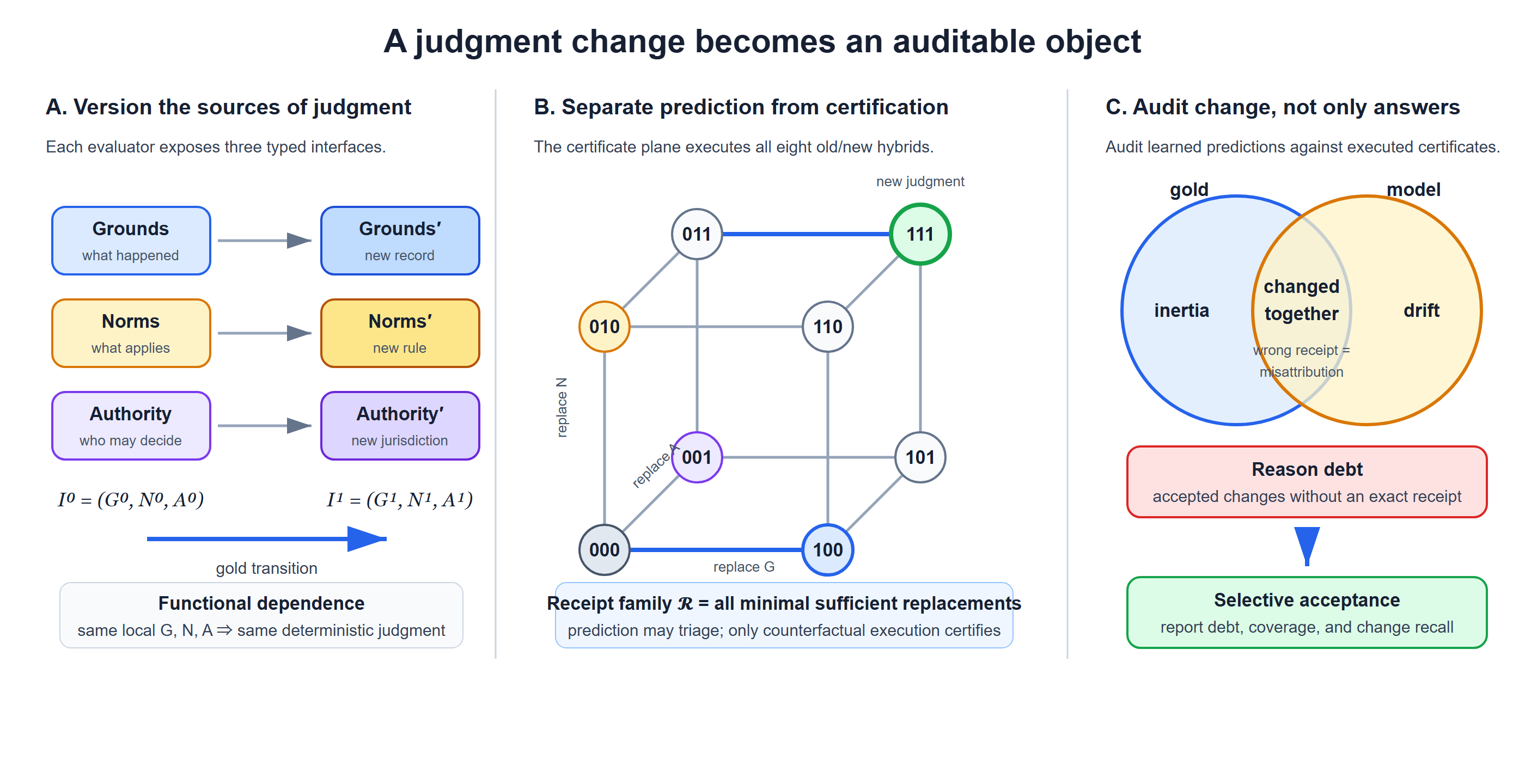}
\caption{The certification workflow. The cube is an executed audit object relative to a declared source interface. A model-generated cube is a prediction of this object, not its certificate.}
\label{fig:1}
\end{figure}

\subsection{Exact enumeration and its cost}

\textbf{Proposition 2 (existence and exact enumeration).} If \(Q_x(\varnothing)\ne Q_x(\mathcal K)\), then \(\mathcal R_x\ne\varnothing\). Querying every hybrid and keeping exactly the sufficient subsets with no sufficient strict subset returns \(\mathcal R_x\).

\emph{Proof.} \(\mathcal K\) is sufficient, so the finite family of sufficient subsets is nonempty and has minimal members. Every kept subset is sufficient and minimal by construction; every receipt is queried and survives because it has no sufficient strict subset. \(\square\)

\textbf{Proposition 3 (receipt antichain).} With \(k\) source types,

\[
|\mathcal R_x|\le \binom{k}{\lfloor k/2\rfloor}.
\]

\emph{Proof.} No receipt can strictly contain another, so \(\mathcal R_x\) is an antichain in the Boolean lattice, and Sperner's theorem bounds its size (Engel, 1997). \(\square\)

With three sources there are at most three receipts.

\textbf{Proposition 4 (worst-case certification lower bound).} Suppose the endpoint labels are known and differ. With no monotonicity or other structural assumption on the black-box hybrid evaluator, every exact deterministic algorithm that certifies the complete receipt family must query all \(2^k-2\) interior hybrids in the worst case.

\emph{Proof.} Let the endpoints be \(y_0\ne y_1\). An adversary answers \(y_0\) to every interior query. If the algorithm stops with subset \(S\) unqueried, two completions match the transcript: one where every interior cell is \(y_0\), making \(\mathcal K\) the sole receipt, and one where exactly \(S\) is \(y_1\), making \(S\) the sole receipt. The required outputs differ. \(\square\)

The bound concerns exact black-box certification. It does not say a model should be trained to emit every cell; that distinction turns out to matter in the experiments.

\subsection{Prediction is not certification}

A language model can emit a receipt directly, or emit a full cube from which a receipt is derived. Both are predictions, and neither certifies itself. The split is the familiar one between verification by execution and estimation by a learned model.

Let \(\widehat{\mathcal R}_x\) be a model output and \(\mathcal R_x^\star\) the family obtained by executing a trusted reference adjudicator over the declared hybrids. Exact predictive accuracy is \(\mathbf 1[\widehat{\mathcal R}_x=\mathcal R_x^\star]\). A \emph{receipt certificate} is the executed table \(\{(S,Q_x^\star(S)):S\subseteq\mathcal K\}\) with the deterministic minimality computation and content identifiers binding case, adjudicator, and the two source states. A predicted cube \(\widehat Q_x\) can force internal consistency between cells and derived receipt; it cannot supply external correctness.

So we keep two words apart:

\begin{quote}

\itshape \textbf{A receipt can be predicted from a description; it is certified only by executing the relevant counterfactuals.}

\end{quote}

The lower bound explains why the full cube is valuable for audit. It promises nothing about whether full-cube supervision is a statistically easier or more robust training target.

\subsection{Paired consistency under controlled transformations}

Some prompt transformations preserve the underlying transition; others transform it in a known way. Let \(z(x)\) be the audited output, endpoints, receipt, or cube, and let \(\tau\) be a transformation with known output relation \(H_\tau\). In software-testing terms \(H_\tau\) is a metamorphic relation (Chen et al., 1998; Segura et al., 2016); in NLP terms the identity-relation cases are behavioral invariance tests (Ribeiro et al., 2020). The \emph{paired-consistency rate} of a predictor is

\[
\operatorname{PC}(\tau)=
\frac{1}{|\mathcal X|}
\sum_{x\in\mathcal X}
\mathbf 1\!\left[
H_\tau\!\left(\widehat z(x),\widehat z(\tau x)\right)
\right].
\]

For a source-order permutation or an irrelevant distractor, \(H_\tau\) demands identical endpoints and receipts. For an inverse transition, it demands that the endpoints swap. A transformation can also specify a receipt relation when one is well defined.

The metric reads nothing inside the model. It tests a behavioral consequence of the declared semantics: if both predictions were exactly right, the relation would hold, so a violated relation proves at least one of the pair wrong. Unpaired accuracy cannot see this, because it never asks whether two predictions move together as required.

\textbf{Proposition 5 (consistency error bound).} Suppose gold outputs satisfy \(H_\tau(z^\star(x),z^\star(\tau x))\). Let

\[
e_\tau(x)=\frac{1}{2}\left(
\mathbf 1[\widehat z(x)\ne z^\star(x)]
+\mathbf 1[\widehat z(\tau x)\ne z^\star(\tau x)]
\right).
\]

Then

\[
\frac{1}{|\mathcal X|}\sum_{x\in\mathcal X}e_\tau(x)
\ge \frac{1-\operatorname{PC}(\tau)}{2}.
\]

\emph{Proof.} If the predicted pair violates \(H_\tau\) while the gold pair satisfies it, both predictions cannot be correct, so \(2e_\tau(x)\ge \mathbf 1[\neg H_\tau(\widehat z(x),\widehat z(\tau x))]\) pointwise. Average. \(\square\)

The bound needs no gold labels for transformed items, no independence, and no model of the error mechanism. A failed relation has a direct reading: at least one member of that pair is wrong. The bound stays silent on how the errors divide between the original and transformed prompts; it is only a floor.

\section{Failure accounting}

Let \(\widehat y_x^0\) and \(\widehat y_x^1\) be the predicted old and revised verdicts. The gold change set and the model change set are

\[
\mathcal A=\{x:Q_x^\star(\varnothing)\ne Q_x^\star(\mathcal K)\},
\qquad
\mathcal M=\{x:\widehat y_x^0\ne\widehat y_x^1\}.
\]

Three failures are distinct, and the first two are ordinary change-detection errors under short names:

\begin{samepage}

\begin{itemize}

\item \emph{drift}, \(\mathcal M\setminus\mathcal A\): the model changes when the gold verdict is stable (a false-positive change);

\item \emph{inertia}, \(\mathcal A\setminus\mathcal M\): the model fails to change when the gold verdict changes (a false-negative change);

\item \emph{misattribution}: on a gold change, the predicted receipt differs from the exact receipt.

\end{itemize}

\end{samepage}

Misattribution can survive perfect endpoints, so the experiments also report receipt error conditional on both endpoint verdicts being correct.

For an accepted set of predicted changes \(\mathcal C\subseteq\mathcal M\), \emph{reason debt} is

\[
\operatorname{Debt}(\mathcal C)=
1-\frac{
\left|\{x\in\mathcal C:
\widehat{\mathcal R}_x=\mathcal R_x^\star\}\right|
}{|\mathcal C|}.
\]

In selective-prediction terms, debt is the selective risk of the attribution task (Geifman and El-Yaniv, 2019). It is undefined on an empty accepted set and must be reported with accepted coverage and gold-change recall; otherwise abstaining on every hard case simulates perfect accounting while entrenching inertia.

The change-set decomposition gives one identity worth stating. An updated evaluator that matches the gold revised evaluator on \(\mathcal A\) and preserves the gold old verdict outside \(\mathcal A\) matches the gold revised evaluator everywhere: inside \(\mathcal A\) by the first condition, outside because old and revised gold verdicts coincide there. Average accuracy can trade the two regions against each other; the identity cannot.

\section{ReasonBench}

ReasonBench is built so that every candidate explanation can be executed. Each case ships executable old and revised evaluator states; the benchmark runs all eight hybrids and stores the exact receipt family. Two strata contribute complementary weaknesses.

\subsection{Organizational policy worlds}

The organizational stratum starts from 45 audited clauses in the public customer-service environments of \(\tau^2\)-Bench (Barres et al., 2025), 15 each from airline, retail, and telecom. Each clause is translated into a typed, executable policy rule. Cases carry structured grounds and an authority scope over roles, regions, products, or channels. Mutations revise one or more sources. Verdicts are \nolinkurl{approve}, \nolinkurl{deny}, and \nolinkurl{refer}.

The translation is deliberately small: it keeps recognizable policy structure while making every counterfactual executable, and it does not claim to simulate a firm. The source commit, clause manifest, translations, and checksums are in the datasheet. The 45-clause inventory and its clause-level split assignment were reused from an earlier public-data construction; no model result from this project informed the split. All states, scopes, cubes, receipts, prompts, and labels were built fresh.

\subsection{Logical rule worlds}

The logical stratum uses the public RuleTaker generator and grammar as a source specification (Clark, Tafjord, and Richardson, 2020). We independently sample 800 compatible positive-rule worlds and run forward chaining; no original labels are copied. Facts serve as grounds, implication rules as norms, and an explicit mask, added by ReasonBench, determines which rules are authoritative for the query.

Logical worlds contribute many independent units and controlled proof depth. Their templated language invites shortcut learning (Geirhos et al., 2020), which is why alternate renderings and source-order controls belong to the core test.

\subsection{Cases, splits, and controls}

Source clauses and worlds are assigned to splits before any prompt is rendered. The locked test, a frozen held-out split fixed before any model output was inspected, contains 2,400 examples from 102 independent units: six organizational clauses and 96 logical worlds. Rendered prompts are never treated as independent observations.

\begin{table}[!htbp]
\centering
\small
\setlength{\tabcolsep}{4pt}
\resizebox{\textwidth}{!}{%
\begin{tabular}{lrrrrrr}
\toprule
Stratum & Independent units & Main cases & Train & Calibration & Locked & Shift-only \\
\midrule
Organizational policy & 45 clauses & 4,320 & 2,592 & 576 & 576 & 576 \\
Logical rules & 800 worlds & 15,200 & 9,120 & 1,520 & 1,824 & 2,736 \\
\textbf{Total} & \textbf{845} & \textbf{19,520} & \textbf{11,712} & \textbf{2,096} & \textbf{2,400} & \textbf{3,312} \\
\bottomrule
\end{tabular}%
}
\end{table}

Among the main cases, 10,888 carry a gold endpoint change and 8,632 are stable. Minimum receipt size is one for 7,366 changed cases, two for 2,693, three for 829; 3,470 cases are receipt-ambiguous. All eight intervention masks occur.

Each locked example generates three paired controls:

\begin{samepage}

\begin{enumerate}

\item an inverse transition, whose endpoints must swap;

\item a permutation of source-section order, whose endpoints and receipts must not move; and

\item an irrelevant-record distractor, whose endpoints and receipts must not move.

\end{enumerate}

\end{samepage}

These 7,200 controls are transformations of locked examples and add no independent observations.

The frozen protocol lists five distribution tests. \nolinkurl{operator_shift} mixes operator and clause changes; \nolinkurl{linguistic_shift} re-renders prompts in an alternate style; \nolinkurl{depth_shift} increases proof depth. \nolinkurl{composition_shift} was found, after freezing, not to withhold intervention combinations, so we relabel it \emph{reserved structural} and do not cite it as compositional evidence. A post-freeze secondary experiment trains on intervention cardinality zero or one and tests on cardinality two or three; that correction was recorded before any learned locked result was opened.

\subsection{Data and run audit}

Before the primary freeze, an independent executor recomputed every cube, receipt, target string, and intervention label: zero label errors, zero source-unit leakage, zero exact or normalized prompt overlap across splits. The control audit reproduced all 7,200 transformations without error.

Sixteen primary runs were required, five seeds each for direct verdict, direct receipt, and judgment cube, plus one descriptive rationale control; all sixteen completed. A separate audit rebuilt every aggregate from saved prediction files and rejected missing, duplicated, malformed, or non-canonical artifacts. Two post-run implementation errors were found in aggregation, not training: a changed-case denominator was initially incremented only after successful parsing, and a heterogeneous CSV table took the first row's fields. Both were logged, fixed, covered by regression tests, and followed by complete reaggregation from raw predictions. Malformed generations count as errors. The final suite holds 25 passing tests.

\section{Experimental design}

\subsection{Learning targets}

All learned methods see identical descriptions of old and revised grounds, norms, and authority. They differ only in the serialized target.

\begin{samepage}

\begin{itemize}

\item \textbf{Direct verdict} emits old and revised verdicts.

\item \textbf{Direct receipt} emits both verdicts plus the complete receipt family.

\item \textbf{Rationale receipt} prefixes a deterministic one-sentence rationale to the direct receipt. One seed, as a descriptive control; open-ended explanation is out of scope.

\item \textbf{Judgment cube} emits all eight verdicts in fixed cell order; endpoints and receipt are derived mechanically.

\end{itemize}

\end{samepage}

Cube outputs are internally coherent by construction, since the derived receipt cannot contradict the predicted cells. Coherence of this kind is a formatting property; whether the cells are right is a separate question.

The frozen primary comparison asked whether cube supervision improves revised-verdict and exact receipt-family accuracy over direct receipt supervision, on the proposed mechanism of denser counterfactual signal. Two implications were frozen with it: any advantage should concentrate on multi-source and receipt-ambiguous cases, and it should survive equivalent presentations. Those implications are what make the hypothesis falsifiable.

\subsection{Model and optimization}

The primary backbone is the pinned Qwen3-1.7B release (Yang et al., 2025) with LoRA adaptation (Hu et al., 2022). Five seeds share identical data, optimization, and decoding.

\begin{table}[!htbp]
\centering
\small
\setlength{\tabcolsep}{4pt}
\resizebox{\textwidth}{!}{%
\begin{tabular}{lr}
\toprule
Setting & Value \\
\midrule
Backbone & Qwen3-1.7B, revision \nolinkurl{70d244cc86ccca08cf5af4e1e306ecf908b1ad5e} \\
Adaptation & LoRA rank 16, alpha 32, dropout 0.05 on attention and MLP projections \\
Unique examples per primary run & 5,184 after equalizing the two strata \\
Optimization & 500 steps; batch 2; accumulation 16; cosine schedule; 3\% warmup \\
Learning rate / weight decay & \(10^{-4}\) / \(10^{-2}\) \\
Context and precision & 1,024 tokens; bfloat16 \\
Seeds & 11, 23, 37, 53, 71 \\
Decoding & Greedy; strict full-string parser \\
\bottomrule
\end{tabular}%
}
\end{table}

The full unbalanced training split holds 11,712 examples. Character TF-IDF and majority baselines use that full split as leakage and surface-predictability diagnostics; they are not compute-matched competitors. A 100-step format-only calibration before freezing produced 256 valid outputs per primary serialization; no locked semantic outcome was opened during it.

\subsection{Outcomes and inference}

Co-primary outcomes: revised-verdict accuracy and exact receipt-family accuracy. Secondary outcomes: endpoint-pair accuracy, cube exact match, drift, inertia, receipt error conditional on correct endpoints, paired-consistency rates, and reason debt under selective acceptance. Stable endpoints have an empty receipt family, so exact receipt accuracy is also reported on gold changes alone; every gold-change example enters that denominator, including malformed generations.

Uncertainty is clustered by independent clause or world. Paired method differences use 10,000 hierarchical bootstrap replicates, resampling the five paired seeds and the independent units with replacement while preserving method pairing within each sampled seed and example. Holm adjustment covers the two combined locked-test co-primary comparisons; strata, shifts, and diagnostic subgroups are reported with intervals but sit outside the confirmatory family. Tables show means with seed standard deviations; effects are percentage points with 95\% cluster-bootstrap intervals.

Selective thresholds are fitted on calibration data at accepted-change coverage targets of 20\% to 100\%. Locked-test reason debt is reported with realized coverage and the recall of correctly accepted gold changes.

The protocol and analysis code were frozen under SHA-256 \nolinkurl{e0a10b800ff7ee3626a3eaeae21fd2b1a708684e779ee0cbda380fdf2345a93e}. This is an internal content-addressed freeze, not third-party preregistration. The true composition holdout and the smaller-backbone replication were added after the split-name audit and are labeled secondary throughout; their changed-only and stratum intervals were added after the combined composition metrics were seen and are descriptive.

\section{Results}

\subsection{Surface diagnostics do not solve the locked test}

The majority diagnostic reaches 53.8\% revised-verdict and 44.2\% exact receipt accuracy on the locked test. Character TF-IDF reaches 90.0\% and 61.7\% despite using more training data than the neural runs. The locked split therefore contains exploitable lexical signal, but lexical prediction does not approach the learned receipt results. Analytic expectations are 22.6\% for a train-prior random receipt and 7.7\% for a uniform draw over the observed receipt support.

\begin{table}[!htbp]
\centering
\caption{Surface and chance diagnostics on the locked test. The character model and majority target use the full unbalanced training split; the random rows are analytic expectations.}
\label{tab:1}
\small
\setlength{\tabcolsep}{4pt}
\resizebox{\textwidth}{!}{%
\begin{tabular}{lrr}
\toprule
Diagnostic & Revised-verdict accuracy & Exact receipt-family accuracy \\
\midrule
Majority target & 53.8\% & 44.2\% \\
Character TF-IDF & 90.0\% & 61.7\% \\
Train-prior random receipt & — & 22.6\% \\
Uniform observed-support receipt & — & 7.7\% \\
\bottomrule
\end{tabular}%
}
\end{table}

These diagnostics rule out a label-frequency account. They do not establish reasoning; the paired controls below carry that burden.

\subsection{Full-cube supervision does not improve prediction}

Every primary method is accurate on the ordinary locked test. The strongest learned receipt target turns out to be direct receipt.

\begin{table}[!htbp]
\centering
\caption{Locked-test outcomes. Parentheses give seed standard deviation in percentage points. The rationale row is descriptive (one seed).}
\label{tab:2}
\small
\setlength{\tabcolsep}{4pt}
\resizebox{\textwidth}{!}{%
\begin{tabular}{lrrr}
\toprule
Method & Revised-verdict accuracy & Exact receipt-family accuracy & Seeds \\
\midrule
Direct verdict & 99.10\% (0.30) & — & 5 \\
Direct receipt & \textbf{99.27\% (0.05)} & \textbf{98.41\% (0.06)} & 5 \\
Judgment cube & 97.42\% (0.33) & 96.99\% (0.44) & 5 \\
Rationale receipt & 98.29\% & 97.75\% & 1 \\
\bottomrule
\end{tabular}%
}
\end{table}

\begin{figure}[tbp]
\centering
\includegraphics[width=\textwidth]{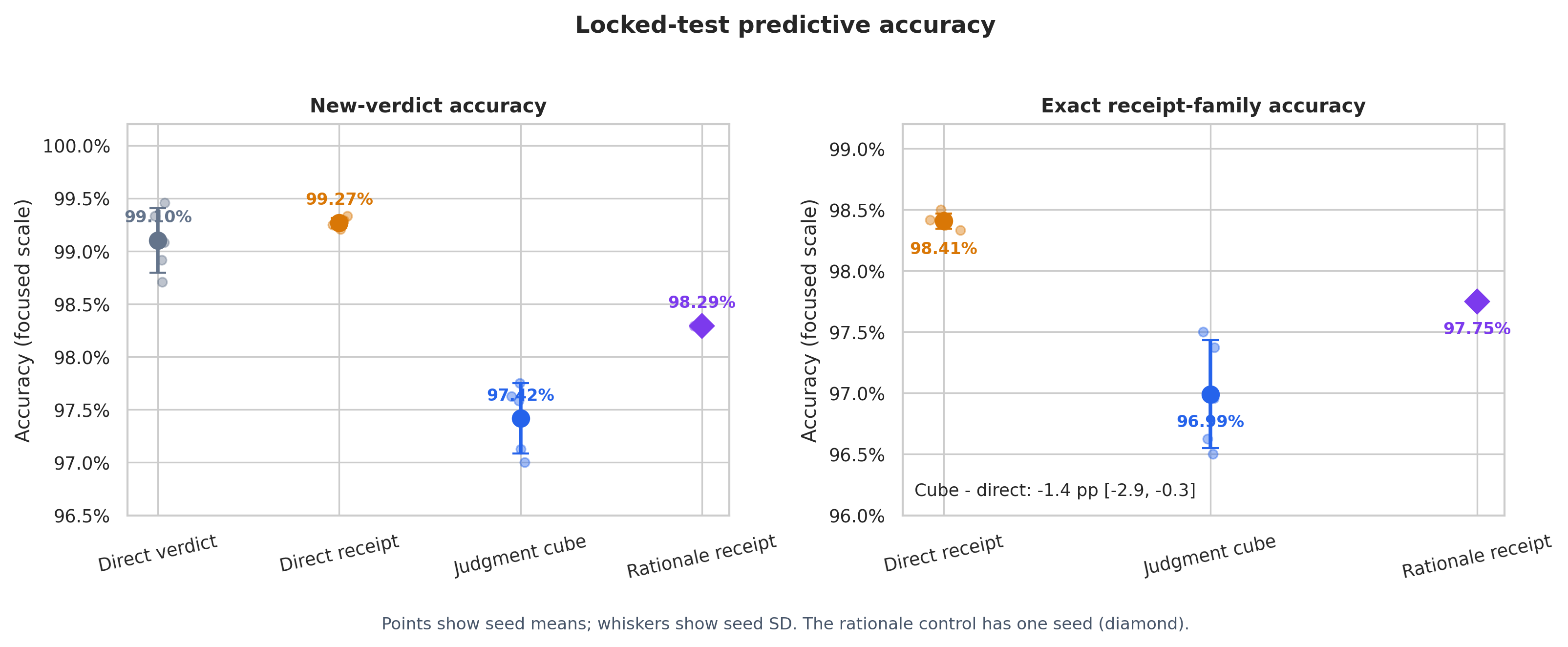}
\caption{Locked-test predictive accuracy. Axes are focused and stated, because zero-based bars would conceal the observed differences.}
\label{fig:2}
\end{figure}

The paired cube-minus-direct effect on exact receipt accuracy is \(-1.42\) percentage points, 95\% interval \([-2.87, -0.30]\), Holm-adjusted \(p=.002\). On revised-verdict accuracy the effect is \(-1.85\) points \([-4.01, -0.23]\), Holm-adjusted \(p=.0036\). The frozen superiority claim is rejected.

The aggregate needs one qualification. On the logical locked stratum the methods are nearly tied in receipt accuracy (\(-0.07\) points \([-0.19, 0.00]\)); the loss concentrates in the six organizational locked clauses, where the effect is \(-5.69\) points \([-10.59, -2.19]\). Six clauses can flag a problem; they cannot estimate how organizational policies behave in general.

The result separates internal coherence from external accuracy. Both methods emit almost perfectly valid formats, and the cube's receipt is consistent with its own cells by construction; the cells are simply wrong more often. Structural discipline does not offset the harder prediction target.

\subsection{Correct verdicts can carry a wrong receipt}

On the 1,340 gold changes per locked seed, direct receipt reaches 97.30\% receipt accuracy, with revised-verdict accuracy 1.54 points higher \([0.24, 3.42]\). The cube's gap is 0.90 points \([-0.62, 2.45]\).

The stricter test conditions on changed cases with both endpoints correct. There, direct prediction still assigns a wrong receipt in 1.66\% of eligible seed-example observations \([0.26\%, 3.61\%]\), the cube in 1.85\% \([0.47\%, 3.57\%]\). Getting the endpoints right does not entail getting the attribution right.

\begin{table}[!htbp]
\centering
\caption{Verdict–receipt separation. Intervals use the paired hierarchical bootstrap over seeds and source units.}
\label{tab:3}
\small
\setlength{\tabcolsep}{4pt}
\resizebox{\textwidth}{!}{%
\begin{tabular}{lrr}
\toprule
Diagnostic on locked gold changes & Direct receipt & Judgment cube \\
\midrule
Exact receipt accuracy & 97.30\% & 95.54\% \\
Revised-verdict minus receipt accuracy & 1.54 pp [0.24, 3.42] & 0.90 pp [-0.62, 2.45] \\
Wrong receipt given both endpoints correct & 1.66\% [0.26, 3.61] & 1.85\% [0.47, 3.57] \\
Eligible seed-example observations & 6,609 & 6,337 \\
\bottomrule
\end{tabular}%
}
\end{table}

Change-set errors are small but present. Direct receipt shows mean drift 0.08\% and inertia 0.31\% over all locked cases; the cube shows 0.52\% and 1.07\%; direct verdict, which has no receipt to score, shows 0.07\% and 0.38\%.

\begin{figure}[tbp]
\centering
\includegraphics[width=\textwidth]{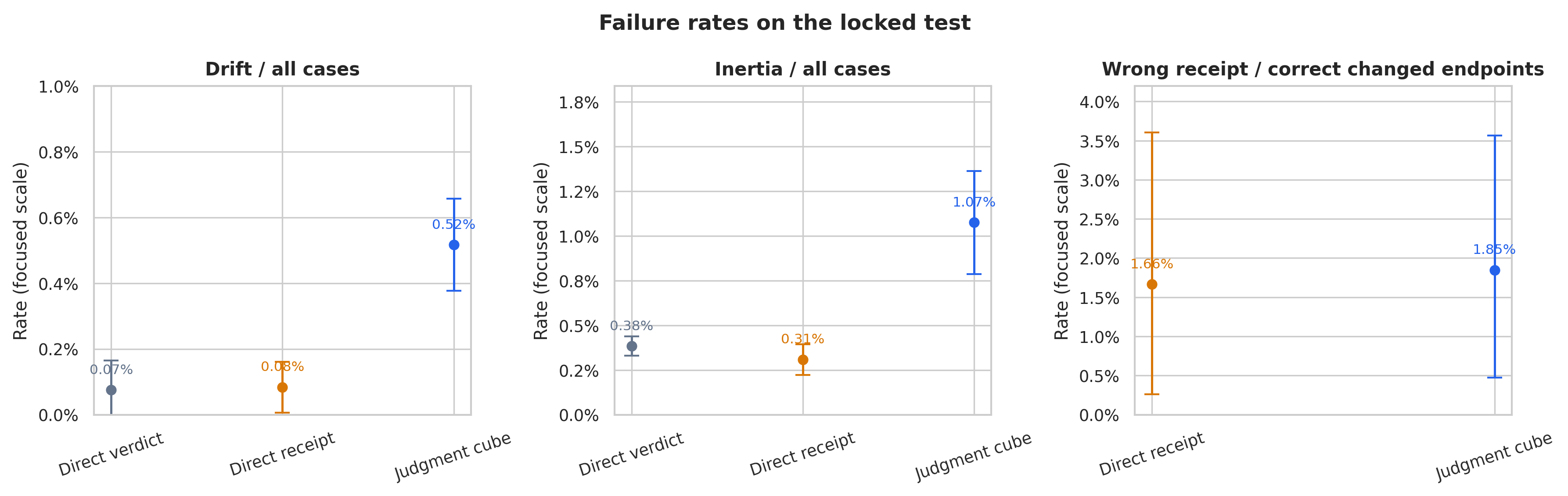}
\caption{Failure rates on the locked test. Drift and inertia whiskers show seed standard deviation; conditional receipt-error whiskers are 95\% cluster-bootstrap intervals.}
\label{fig:3}
\end{figure}

\subsection{More structure does not reveal a cube advantage}

Direct prediction stays ahead at minimum receipt sizes one and two; at size three the methods tie at the displayed precision. Receipt-ambiguous cases are not harder than the rest of the locked test here, and cube supervision does not improve them.

\begin{table}[!htbp]
\centering
\caption{Exact receipt accuracy by structure. Intervals are shown for the prespecified paired subgroup comparisons. Rows differ in structure and source composition at once; they are not isolated causal effects.}
\label{tab:4}
\small
\setlength{\tabcolsep}{4pt}
\resizebox{\textwidth}{!}{%
\begin{tabular}{lrrrr}
\toprule
Locked subgroup & Cases per seed & Direct receipt & Judgment cube & Cube minus direct \\
\midrule
Minimum size 1 & 904 & 98.21\% & 96.17\% & -2.04 pp [-4.79, -0.27] \\
Minimum size 2 & 335 & 95.40\% & 93.85\% & -1.55 pp [-4.06, 0.24] \\
Minimum size 3 & 101 & 95.45\% & 95.45\% & 0.00 pp [-0.99, 1.19] \\
Not receipt-ambiguous (includes stable cases) & 1,980 & 98.38\% & 96.81\% & — \\
Multiple minimal receipts & 420 & 98.52\% & 97.86\% & -0.67 pp [-1.87, 0.29] \\
\bottomrule
\end{tabular}%
}
\end{table}

\begin{figure}[tbp]
\centering
\includegraphics[width=\textwidth]{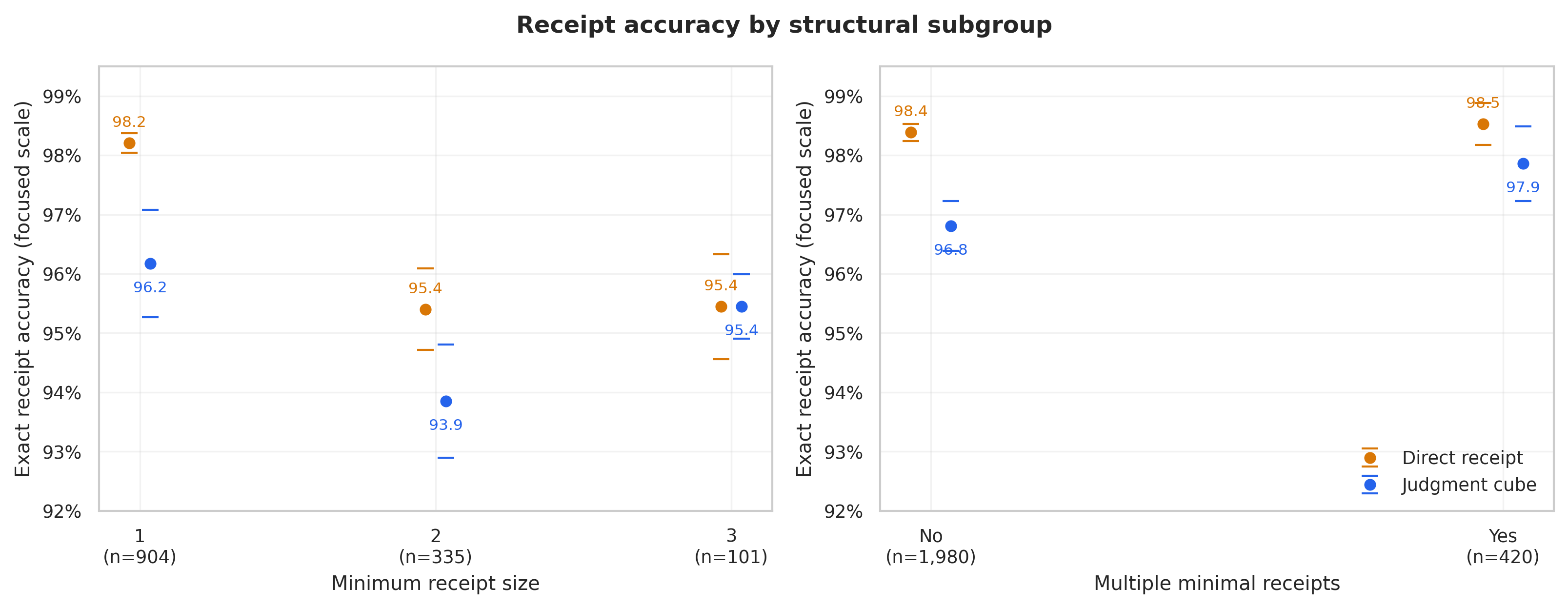}
\caption{Structural subgroups. Whiskers show seed standard deviation.}
\label{fig:4}
\end{figure}

This rejects the frozen mechanism as stated. If dense counterfactual supervision taught reusable receipt structure, the relative gain should surface on multi-source or multiple-receipt cases. It does not.

\subsection{Ordinary shifts and consistency controls tell different stories}

Distribution shift costs receipt accuracy, with the size depending on shift and target.

\begin{table}[!htbp]
\centering
\caption{Exact receipt accuracy under distribution shift. Effects are percentage points from paired hierarchical bootstrap estimates. The reserved structural split is not an unseen-composition test.}
\label{tab:5}
\small
\setlength{\tabcolsep}{4pt}
\resizebox{\textwidth}{!}{%
\begin{tabular}{lrrrr}
\toprule
Split & Cases per seed & Direct receipt & Judgment cube & Cube minus direct (95\% CI) \\
\midrule
Locked & 2,400 & 98.41\% & 96.99\% & -1.42 [-2.87, -0.30] \\
Mixed operator & 288 & 81.81\% & 78.61\% & -3.19 [-8.61, 2.36] \\
Language & 1,200 & 80.50\% & 68.40\% & -12.10 [-16.35, -7.50] \\
Reserved structural & 912 & 99.41\% & 96.27\% & -3.14 [-5.92, -0.86] \\
Proof depth & 912 & 96.73\% & 82.19\% & -14.54 [-19.34, -8.00] \\
\bottomrule
\end{tabular}%
}
\end{table}

The mixed-operator split is not uniformly novel. Its genuinely held-out exception-rule subset has 192 cases from two organizational clauses; direct receipt reaches 80.52\%, the cube 79.58\%, effect \(-0.94\) points \([-6.88, 3.75]\). Two source units cannot support a rule-family claim, so the interval is descriptive.

The paired controls are the harder examination, because each specifies the relation that must hold between predictions on the parent and transformed prompts.

\begin{table}[!htbp]
\centering
\caption{Paired-consistency rates. Each relation is evaluated on 2,400 paired locked transformations per seed. Controls are transformations, not an additional independent sample.}
\label{tab:6}
\small
\setlength{\tabcolsep}{4pt}
\resizebox{\textwidth}{!}{%
\begin{tabular}{lrrr}
\toprule
Paired control & Required relation & Direct receipt & Judgment cube \\
\midrule
Inverse transition & Endpoint pair swaps & 47.22\% & 42.97\% \\
Source-order permutation & Receipt unchanged & 54.77\% & 49.18\% \\
Irrelevant distractor & Receipt unchanged & 98.63\% & 93.43\% \\
\bottomrule
\end{tabular}%
}
\end{table}

\begin{figure}[tbp]
\centering
\includegraphics[width=\textwidth]{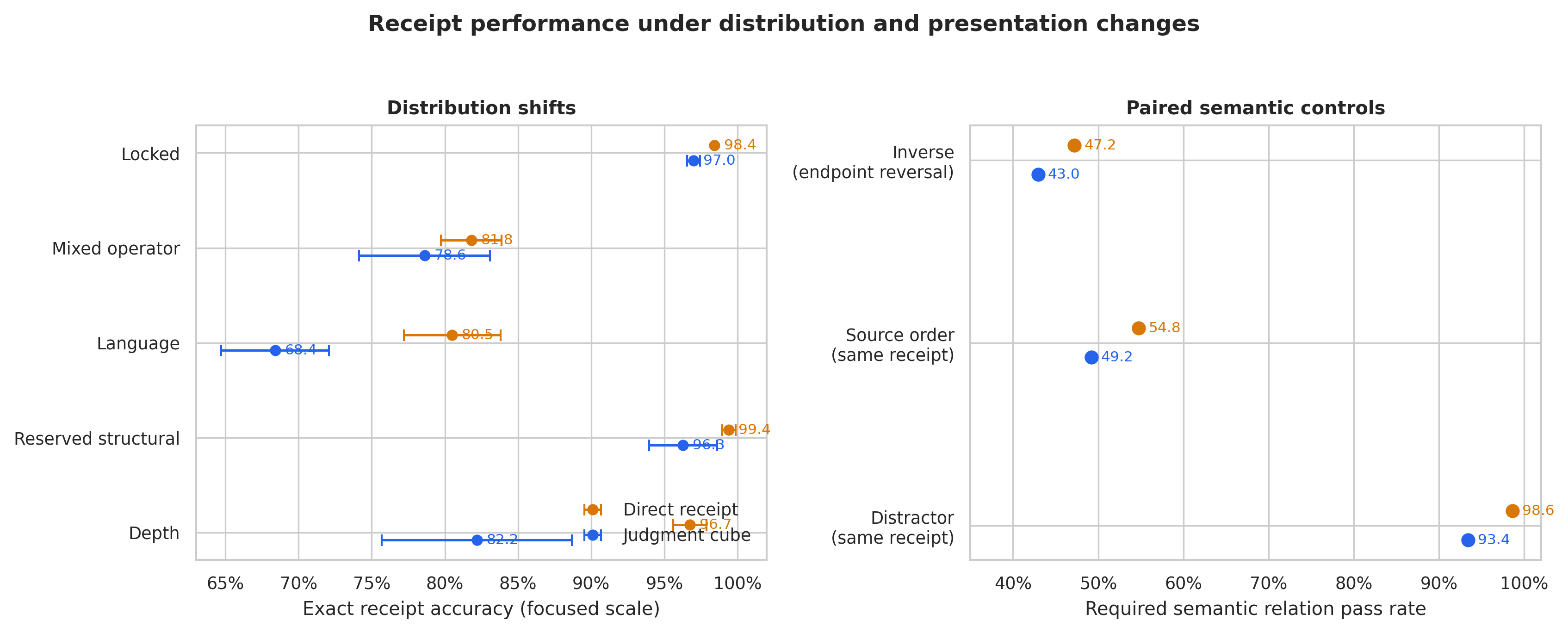}
\caption{Distribution versus presentation changes. Left: absolute receipt accuracy with seed-SD whiskers. Right: the semantic relation required by each paired transformation.}
\label{fig:5}
\end{figure}

The distractor result shows the models can ignore one added irrelevant record. The other two controls land differently. Reordering semantically typed sections destroys nearly half of the required receipt relations even though the gold receipt is unchanged, and reversing the transition fails more than half the time. By Proposition 5, a 54.77\% order-consistency rate already forces a mean pairwise error of at least 22.6\% on those pairs, whatever the unpaired scores say. Near-ceiling locked accuracy is therefore not evidence of a presentation-independent account of the revision. Section~\ref{sec:permaug} asks whether training-time augmentation repairs these failures.

\subsection{Permutation-augmented training (post-hoc)}
\label{sec:permaug}

A reader of the frozen study can name three routes by which the negative result might not generalize: larger scale, constrained decoding, and permutation augmentation. This post-hoc study takes the third at the primary scale. Each training example keeps its original prompt with probability one half; otherwise its three source sections are regrouped in one of the six component orders, sampled uniformly per example. Labels, targets, splits, and every other setting stay at their frozen values: pinned backbone, LoRA rank 16, 500 steps, greedy decoding, strict parser. Two targets, three seeds (11, 37, 71). Every evaluation uses the untouched frozen splits and controls. The frozen columns of Table~\ref{tab:permaug} restate the matching three-seed subset of the primary runs, so all contrasts are paired, and intervals use the same seed-and-unit bootstrap as the rest of the paper.

\begin{table}[!htbp]
\centering
\caption{Post-hoc permutation-augmented training. Frozen columns restate the seed-11/37/71 subset of the primary runs, so columns are paired within each target; parentheses give seed standard deviation in percentage points. Consistency rows are paired-consistency rates against the locked parents; indented rows split the measure by source stratum. The augmentation targets only the order transformation.}
\label{tab:permaug}
\small
\setlength{\tabcolsep}{4pt}
\resizebox{\textwidth}{!}{%
\begin{tabular}{lrrrr}
\toprule
 & \multicolumn{2}{c}{Direct receipt} & \multicolumn{2}{c}{Judgment cube} \\
\cmidrule(lr){2-3}\cmidrule(lr){4-5}
Metric & Frozen & Augmented & Frozen & Augmented \\
\midrule
Locked: receipt, all cases & 98.39\% (0.05) & 98.07\% (0.25) & 96.99\% (0.38) & 92.36\% (1.86) \\
\quad organizational (6 clauses) & 93.29\% (0.20) & 92.25\% (1.12) & 87.67\% (1.39) & 68.52\% (8.18) \\
\quad logical (96 worlds) & 100.00\% (0.00) & 99.91\% (0.06) & 99.93\% (0.06) & 99.89\% (0.15) \\
Locked: receipt, changes only & 97.31\% (0.07) & 96.62\% (0.44) & 95.50\% (0.35) & 88.26\% (3.19) \\
Locked: revised verdict & 99.26\% (0.02) & 99.25\% (0.04) & 97.40\% (0.35) & 92.75\% (1.70) \\
Order consistency & 55.85\% (2.37) & 96.61\% (0.44) & 49.22\% (1.91) & 89.89\% (0.63) \\
\quad organizational (6 clauses) & 49.83\% (1.71) & 87.79\% (2.71) & 51.27\% (2.17) & 67.94\% (1.30) \\
\quad logical (96 worlds) & 57.75\% (2.66) & 99.40\% (0.67) & 48.57\% (3.20) & 96.82\% (1.22) \\
Inverse consistency & 46.60\% (1.11) & 46.96\% (0.29) & 43.01\% (1.09) & 46.93\% (1.81) \\
Distractor consistency & 98.36\% (0.72) & 98.67\% (0.14) & 94.11\% (4.78) & 91.69\% (5.26) \\
Language shift: receipt & 79.89\% (2.37) & 87.31\% (5.22) & 67.31\% (3.72) & 82.81\% (3.34) \\
Proof depth: receipt & 96.86\% (1.21) & 95.69\% (2.69) & 84.32\% (8.19) & 86.40\% (2.84) \\
\bottomrule
\end{tabular}%
}
\end{table}

Augmentation largely restores the relation it targets, with a clear split between strata. Order consistency rises by +40.76 points \([37.33, 44.01]\) for direct receipt and +40.67 \([35.97, 46.04]\) for the cube. On the logical stratum the repair is nearly complete (99.40\% direct, 96.82\% cube); on the six organizational clauses it is partial (87.79\% and 67.94\%), and the augmented cube still violates the required relation on roughly a third of organizational pairs. By Proposition 5, the implied lower bound on mean pairwise error falls from 22.1\% to 1.7\% for direct prediction. Part of the gain could be simple familiarity, since the component-grouped layouts enter the augmented training distribution by construction. The linguistic shift says otherwise: on that rendering, which training never shows, receipt accuracy rises +7.42 points \([1.25, 15.65]\) for direct and +15.50 \([9.60, 20.27]\) for the cube. The inverse relation, which the augmentation does not touch, barely moves for direct receipt (+0.36 \([-0.94, +1.91]\)); the cube's combined shift (+3.92 \([0.53, 8.06]\)) comes from the six organizational clauses. Near 47\%, both models still fail to swap endpoints more often than not.

The costs fall unevenly across targets and strata. Direct receipt pays nothing detectable on the frozen rendering overall (\(-0.32\) points \([-0.73, +0.09]\)), a small real cost on changed cases (\(-0.70\) \([-1.42, -0.03]\)), and no significant cost within either stratum alone (organizational \(-1.04\) \([-2.43, +0.69]\); logical \(-0.09\) \([-0.22, +0.00]\)). The cube's loss sits almost entirely in the organizational stratum: \(-19.16\) points of organizational locked receipt accuracy \([-27.66, -10.42]\), against \(-0.04\) \([-0.29, +0.15]\) on the logical stratum. The combined cube-minus-direct gap accordingly widens from \(-1.40\) \([-2.92, -0.22]\) in the matched frozen subset to \(-5.71\) \([-9.75, -1.94]\), and under augmentation that gap is \(-23.73\) \([-29.75, -15.28]\) on the six organizational clauses versus \(-0.02\) \([-0.27, +0.20]\) on the 96 logical worlds. Six clauses cannot support a claim about organizational policy in general; what they show is that the joint target degrades most exactly where it was already weakest.

Three seeds, one augmentation rate, and a six-clause organizational stratum make this an exploratory, secondary result. Within those limits, two observations stand. First, the near-50\% order-consistency rates of the frozen models were largely a property of the training distribution: a cheap augmentation moves the logical rate from 57.75\% to 99.40\%, and the organizational rate by roughly 38 points while leaving it well short of repair. That is the strongest argument for measuring consistency at all, because the frozen models' near-ceiling accuracy said nothing about it. The improvement is also relation-specific; the untargeted inverse relation stays broken for both targets, so passing one consistency control certifies nothing about the next. Second, in this setting the harder joint target suffers most from the broader training distribution, and it does so where it was already weakest. The frozen study's conclusion, that a more structured target is not automatically a more robust learner, shows up again under a training distribution built to help robustness. Whether it would survive other serializations, per-cell supervision, or constrained decoding remains untested.

\subsection{Selective acceptance lowers debt by reviewing more changes}

Calibrated selective acceptance reduces reason debt only by withholding part of the change set. At the nominal 80\% operating point, direct receipt realizes 81.97\% accepted-change coverage, 0.27\% reason debt, and 81.61\% recall of correctly accepted gold changes; the cube realizes 85.54\%, 0.16\%, and 84.30\%. Near full coverage, debt rises to 2.26\% and 3.48\% while correct gold-change recall is 98.60\% and 94.58\%.

\begin{table}[!htbp]
\centering
\caption{Selected locked-test operating points. Thresholds are fitted on calibration data; coverage, debt, and recall are seed means.}
\label{tab:7}
\small
\setlength{\tabcolsep}{4pt}
\resizebox{\textwidth}{!}{%
\begin{tabular}{lrrr}
\toprule
Method & Realized accepted-change coverage & Reason debt & Correct gold-change recall \\
\midrule
Direct receipt, 80\% target & 81.97\% & 0.27\% & 81.61\% \\
Judgment cube, 80\% target & 85.54\% & 0.16\% & 84.30\% \\
Direct receipt, 100\% target & 99.94\% & 2.26\% & 98.60\% \\
Judgment cube, 100\% target & 99.98\% & 3.48\% & 94.58\% \\
\bottomrule
\end{tabular}%
}
\end{table}

\begin{figure}[tbp]
\centering
\includegraphics[width=\textwidth]{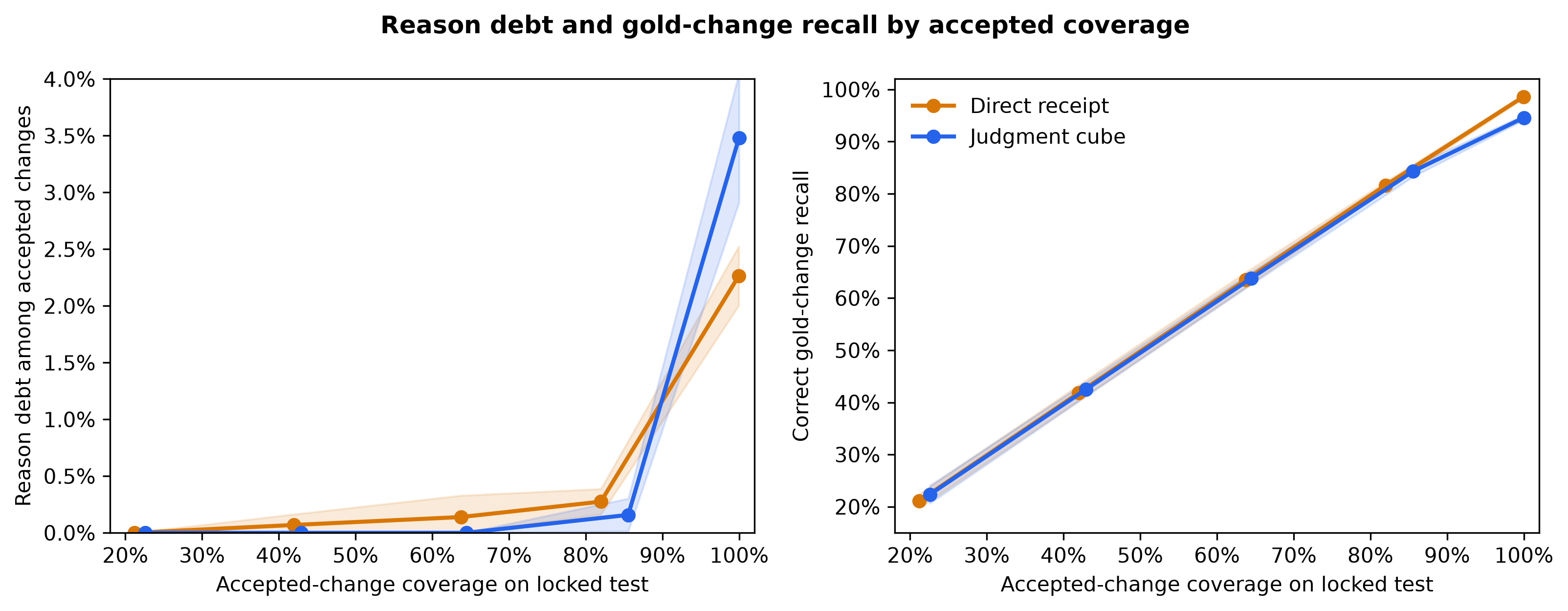}
\caption{Selective accounting. Low debt at partial coverage should be read together with the lost gold-change recall.}
\label{fig:6}
\end{figure}

Selective confidence is operationally useful, but nothing here shows it recognizes a causal failure mode. It shows a calibrated score can concentrate receipt errors. The review burden stays in the accounting.

\subsection{True composition holdout}

The secondary composition experiment trains on intervention cardinality zero or one and evaluates on two or three, with disjoint source units, three seeds per target. The outcome sits far below the locked scores.

\begin{table}[!htbp]
\centering
\caption{Post-freeze true-composition holdout. Parentheses give seed standard deviation; intervals use the same paired source-unit bootstrap as the primary study. Changed-only rows use conditional denominators.}
\label{tab:8}
\small
\setlength{\tabcolsep}{4pt}
\resizebox{\textwidth}{!}{%
\begin{tabular}{lrrrr}
\toprule
Split and metric & Cases per seed & Direct receipt & Judgment cube & Cube minus direct (95\% CI) \\
\midrule
Unseen compositions: revised verdict & 1,440 & 93.75\% (0.91) & 66.94\% (0.50) & -26.81 pp [-30.95, -22.70] \\
Unseen compositions: receipt, all cases & 1,440 & 36.62\% (0.11) & 37.22\% (0.57) & +0.60 pp [-0.37, 1.61] \\
Unseen compositions: receipt, changes only & 968 & 7.16\% (0.06) & 9.57\% (0.70) & +2.41 pp [0.35, 4.50] \\
Unseen compositions plus greater depth: revised verdict & 576 & 99.83\% (0.30) & 59.20\% (3.25) & -40.63 pp [-45.78, -36.57] \\
Unseen compositions plus greater depth: receipt, all cases & 576 & 33.33\% (0.00) & 33.33\% (0.00) & 0.00 pp [0.00, 0.00] \\
Unseen compositions plus greater depth: receipt, changes only & 384 & 0.00\% (0.00) & 0.00\% (0.00) & 0.00 pp [0.00, 0.00] \\
\bottomrule
\end{tabular}%
}
\end{table}

Stable cases carry most of the all-case receipt score: 472 of the 1,440 holdout cases and 192 of the 576 depth cases are stable. On actual changes both targets sit near the floor, and at greater depth neither predicts a single receipt family correctly, while direct prediction keeps 93.75\% and 99.83\% revised-verdict accuracy. Dispositions survive; the account of why they changed does not.

The changed-only comparison needs one more qualification. Its 2.41-point cube advantage comes entirely from 200 organizational changes across six clauses, where direct and cube reach 34.67\% and 46.33\% (stratum effect +11.67 points \([3.76, 16.84]\)); on 768 logical changes across 96 worlds both score zero. The small combined gain coexists with a 26.81-point verdict loss and vanishes in the larger stratum. Cube supervision moves errors around; it does not become a useful composition learner here.

\subsection{Smaller-backbone replication}

The capacity replication repeats the direct-versus-cube comparison with the pinned Qwen3-0.6B backbone on the frozen splits and controls, three seeds, labeled secondary. The locked ordering replicates: direct receipt 98.06\%, cube 96.79\%, effect \(-1.26\) points \([-3.08, -0.04]\); revised-verdict 99.15\% versus 96.76\%, effect \(-2.39\) points \([-4.75, -0.49]\).

\begin{table}[!htbp]
\centering
\caption{Post-freeze Qwen3-0.6B replication. Parentheses give seed standard deviation; effects use paired source-unit bootstrap intervals; changed-only rows show conditional denominators.}
\label{tab:9}
\small
\setlength{\tabcolsep}{4pt}
\resizebox{\textwidth}{!}{%
\begin{tabular}{lrrrr}
\toprule
Split and metric & Cases per seed & Direct receipt & Judgment cube & Cube minus direct (95\% CI) \\
\midrule
Locked: revised verdict & 2,400 & 99.15\% (0.25) & 96.76\% (0.95) & -2.39 pp [-4.75, -0.49] \\
Locked: receipt, all cases & 2,400 & 98.06\% (0.06) & 96.79\% (1.01) & -1.26 pp [-3.08, -0.04] \\
Locked: receipt, changes only & 1,340 & 96.79\% (0.22) & 95.12\% (1.24) & -1.67 pp [-4.36, 0.15] \\
Language shift: revised verdict & 1,200 & 95.44\% (1.18) & 92.97\% (3.00) & -2.47 pp [-8.02, 2.66] \\
Language shift: receipt, all cases & 1,200 & 78.64\% (2.68) & 70.97\% (1.64) & -7.67 pp [-11.09, -3.71] \\
Language shift: receipt, changes only & 662 & 63.24\% (3.98) & 50.15\% (5.39) & -13.09 pp [-18.80, -4.96] \\
Greater depth: revised verdict & 912 & 99.05\% (1.27) & 99.63\% (0.54) & +0.58 pp [-0.77, 2.63] \\
Greater depth: receipt, all cases & 912 & 74.78\% (3.24) & 71.97\% (0.64) & -2.81 pp [-6.58, 0.29] \\
Greater depth: receipt, changes only & 528 & 56.50\% (5.54) & 52.15\% (0.67) & -4.36 pp [-10.80, 0.95] \\
\bottomrule
\end{tabular}%
}
\end{table}

The language rows strengthen the direct-target advantage; the depth interval crosses zero while both methods lose substantial receipt accuracy relative to 1.7B. Capacity matters for deeper rule worlds, but the tested reduction reveals no cube advantage. Format validity is 100\% for both methods on locked, 99.97\% on language shift, and 99.05\% direct versus 100\% cube at depth; formatting does not explain the receipt gaps.

\begin{figure}[tbp]
\centering
\includegraphics[width=\textwidth]{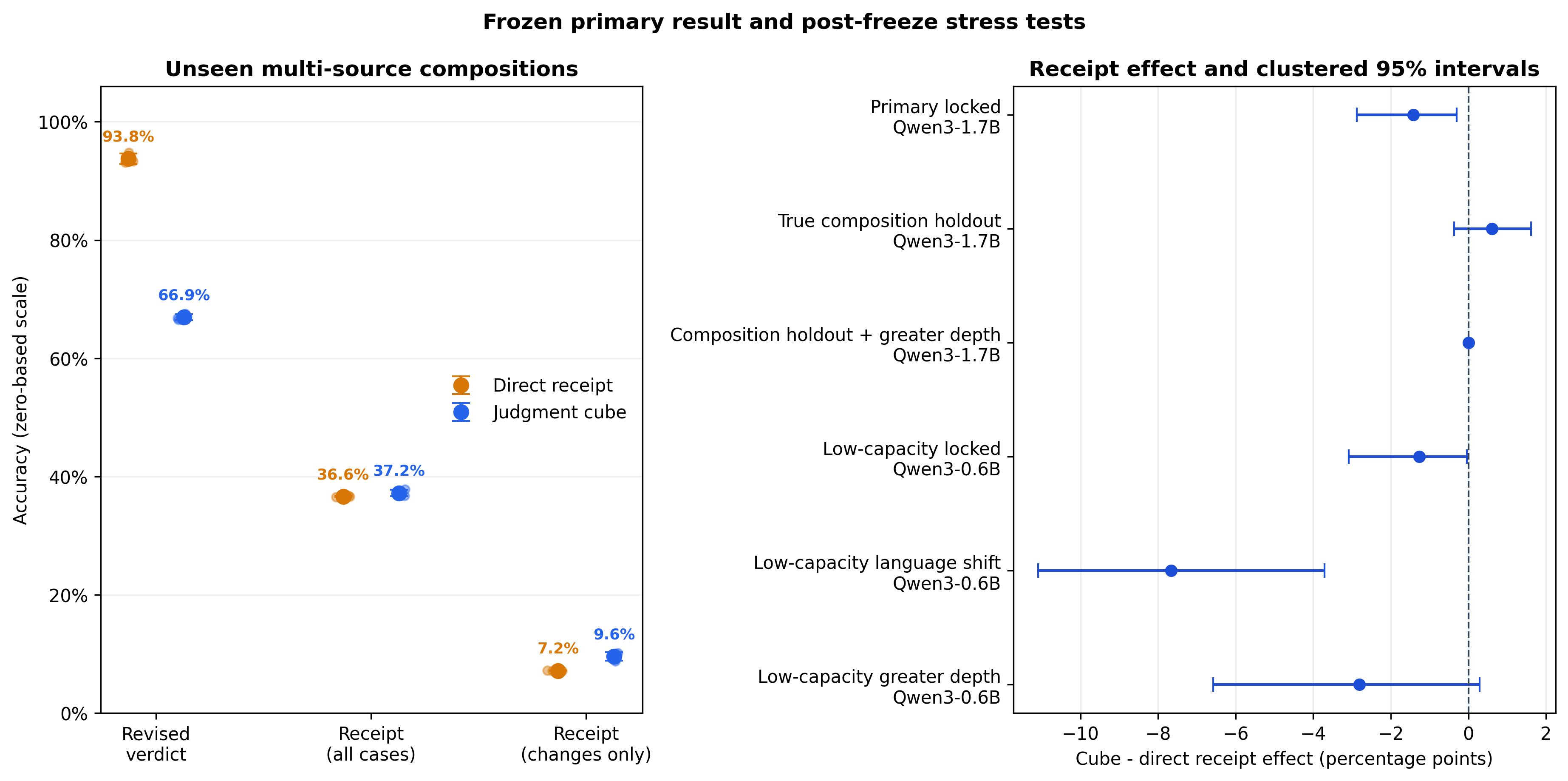}
\caption{Primary and secondary target comparisons. Left: zero-based separation between verdict and changed-case receipt accuracy under unseen source combinations. Right: clustered 95\% intervals; composition and capacity rows are post-freeze secondary evidence.}
\label{fig:7}
\end{figure}

\subsection{Compute}

The primary methods share 17.43 million trainable parameters and a 500-step budget. Mean training time per seed is 53.6 minutes for direct receipt and 53.2 for the cube on the recorded cloud configuration; peak reserved memory is 12.57 and 12.60 GiB. Target length does not explain the cube deficit in the obvious direction: direct receipt generates 14.02 tokens per evaluated example on average, the compact cube serialization 6.70.

\begin{table}[!htbp]
\centering
\caption{Compute and generation cost. Primary rows average five seeds; the rationale row has one. Hardware concurrency makes wall time descriptive rather than platform-independent.}
\label{tab:10}
\small
\setlength{\tabcolsep}{4pt}
\resizebox{\textwidth}{!}{%
\begin{tabular}{lrrrr}
\toprule
Method & Trainable parameters & Mean training time & Peak reserved memory & Generated tokens/example \\
\midrule
Direct verdict & 17.43M & 53.8 min & 12.05 GiB & 7.00 \\
Direct receipt & 17.43M & 53.6 min & 12.57 GiB & 14.02 \\
Judgment cube & 17.43M & 53.2 min & 12.60 GiB & 6.70 \\
Rationale receipt & 17.43M & 52.7 min & 11.43 GiB & 22.22 \\
\bottomrule
\end{tabular}%
}
\end{table}

\section{What the experiment changes}

\subsection{The cube has two roles, and only one is guaranteed}

The theory guarantees an audit property: given a trusted adjudicator and a declared interface, the executed cube identifies the complete minimal receipt family exactly, and it is worst-case query-optimal for unrestricted black-box certification.

The theory guarantees no learning property. A sequence model trained to emit eight cells faces a larger joint prediction problem; any cell error can corrupt the derived receipt, and cells that never touch the endpoints can still break minimality. Direct receipt training compresses the target to the object the metric scores and can exploit benchmark regularities more efficiently. The primary result fits that account.

The apparent paradox is two senses of \emph{stronger}. The full cube is stronger \emph{evidence} for a receipt, since it permits replay and minimality checks. That does not make it the better \emph{statistical target}. The frozen design assumed the senses coincide; the experiment says they need not. Stated in hindsight this is almost expected, which is why we weight the paper toward the failures the comparison exposed rather than the rejected hypothesis itself.

\subsection{High prediction accuracy is compatible with shallow reason learning}

The locked test and the consistency controls ask different questions. The locked test asks whether the model predicts receipts for new source units rendered in a familiar protocol. The order control asks whether the prediction is invariant to an equivalent presentation of the same typed sources. A model can pass the first on stable correlations between section position, amendment form, and labels; passing the second requires more of the task's semantics.

The 98.4\% locked receipt score is real and narrower than it looks: it establishes predictability under the frozen rendering, not a presentation-independent concept of grounds, norm, and authority. The roughly 50\% order-consistency rates are direct evidence against the stronger reading. They are also largely a property of the training distribution. Retraining with permuted source order lifts the combined rate above 96\% for the direct target at no detectable locked cost (\S\ref{sec:permaug}); the six organizational clauses improve less than the logical worlds, and the untargeted inverse relation stays broken either way.

The composition holdout adds a second separation. Direct prediction often preserves the revised verdict on unseen multi-source changes while failing to name their minimal replacements, and here training withheld the intervention combinations themselves; the rendering never changed. The 0.6B replication shows the direct-versus-cube ordering is not an artifact of one backbone; its depth degradation warns against claiming capacity-independence.

This says more than the generic complaint that benchmarks harbor shortcuts. A paired test on the same cases, with a known required relation, attaches the failure to the model–serialization system itself; no appeal to an unobserved deployment distribution is needed.

\subsection{Receipts remain useful after the supervision hypothesis fails}

The negative result does not remove the need for an account of evaluator change; it relocates the account.

For a high-stakes evaluator, the authoritative receipt should come from executing versioned policy components or another trusted adjudication procedure. A small learned model can propose receipts, order reviews, or approximate a costly audit, with its output labeled a prediction and measured against executed certificates. The model then carries an efficient approximation of practice, while the executable interface carries the auditable commitment.

The two roles belong to two planes, a proposer and a checker in verification terms. The \textbf{prediction plane} returns a provisional tuple \((\widehat y^0,\widehat y^1,\widehat{\mathcal R},c)\) with confidence \(c\) for routing and triage. The \textbf{certificate plane} resolves versioned artifacts and executes

\[
\operatorname{Cert}(x;D,I^0,I^1)
=\left(h(x,D,I^0,I^1),Q_x,\mathcal R_x\right),
\]

with \(h\) binding the executed object to case, adjudicator, and source versions. Only the certificate plane discharges reason debt. The prediction plane can cut the number or order of expensive audits, but its coverage, debt, and gold-change recall must be estimated against certificate-plane executions. Our experiments compare a prediction plane against an exact certificate plane; they do not estimate deployment latency or labor savings.

The separation leaves three questions that should not collapse into one score:

\begin{samepage}

\begin{enumerate}

\item \textbf{Fidelity:} does the predicted receipt equal the executed receipt?

\item \textbf{Consistency:} do predictions obey known relations under controlled transformations?

\item \textbf{Certification:} can the claimed account be replayed from the bound case, adjudicator, and source versions?

\end{enumerate}

\end{samepage}

Locked accuracy answers the first on one rendering; paired consistency answers the second; the executed cube with content bindings answers the third. A high value on any one implies neither of the others.

The division bears on specialized application companies directly. Compressing domain experience into weights does not make it governable. A versioned rule, evidence, and authority layer lets an organization separate what it explicitly changed from what the model merely inferred, and the receipt becomes both a filter on model-generated actions and a record for institutional learning.

\subsection{What the failed hypothesis rules out}

The frozen mechanism made three predictions: cube supervision would raise exact receipt accuracy; the gain would grow on multi-source or ambiguous receipts; the gain would survive equivalent presentations. The primary study supports none. Direct receipt wins the locked comparison, no cube advantage appears at higher receipt complexity, and both targets fail order and inverse controls.

The post-freeze composition study narrows rather than reverses the verdict. Cube supervision improves changed-only receipt accuracy by 2.41 points, but both methods stay under 10\%, the gain lives in six organizational clauses, and cube verdict accuracy collapses. Structured supervision can move errors between endpoints and receipts; that is not a recommendation. The augmentation study (\S\ref{sec:permaug}) tests a further escape route: in that setting, an order-diverse training distribution built to help robustness widens the cube's deficit instead of erasing it, with the loss concentrated in the six-clause organizational stratum.

Keeping the failure visible sets the bar for successors. A future method should not be judged on the same locked rendering alone. It should improve paired consistency, true composition, and cross-rendering performance without adding drift or review burden. The benchmark converts a failed architecture claim into that stricter standard.

\section{Related work}

\subsection{Evaluators in the oversight loop}

The motivating deployment is the oversight pipeline: learned judges score model outputs for release, routing, and training feedback (Zheng et al., 2023; Ouyang et al., 2022), and the difficulty of supervising systems at or beyond human evaluation capacity is the scalable-oversight problem (Bowman et al., 2022). Known evaluator pathologies include position and formatting biases (Zheng et al., 2023), sensitivity to where information sits in context (Liu et al., 2024), and shortcut features standing in for the intended rule (Geirhos et al., 2020). Our source-order fragility result is a version-transition analogue of those biases, measured against an executable gold account.

The two-plane architecture, a cheap learned proposer checked by an executed verifier, parallels debate and verification proposals (Irving et al., 2018) and process supervision, which scores intermediate steps rather than final answers alone (Lightman et al., 2024). Making norms an explicit, versioned layer rather than an implicit property of weights is in the spirit of constitutional approaches (Bai et al., 2022), while our authority interface adds the jurisdictional question those approaches leave implicit.

\subsection{Faithful explanations and rationales}

Interpretability distinguishes plausible from faithful explanations (Jacovi and Goldberg, 2020); saliency sanity checks show attractive attributions can be insensitive to model and data (Adebayo et al., 2018); chain-of-thought text can misstate the factors that moved the answer (Turpin et al., 2023; Lanham et al., 2023). ERASER operationalizes comprehensiveness and sufficiency for textual rationales (DeYoung et al., 2020); right-for-the-right-reasons training constrains explanations during learning (Ross, Hughes, and Doshi-Velez, 2017).

Receipts keep the sufficiency demand and change the unit of analysis: the explained object is a transition between evaluator versions, candidate interventions are typed source replacements rather than token subsets, and the complete minimal family is retained. Self-generated counterfactual explanations show a validity–minimality tradeoff (Mayne et al., 2025) and can disagree with the model's own predictions (Dehghanighobadi, Fischer, and Zafar, 2025); executed behave-as-claimed tests compare stated policies with counterfactual behavior (Shi et al., 2025). ReasonBench fixes the intervention set in advance and executes every candidate through a reference adjudicator.

Human explanation is contrastive and selective (Miller, 2019). A receipt is contrastive, asking why the revised verdict rather than the old one, but its selection rule is mechanical minimality, not salience.

\subsection{Sufficient reasons, causality, and provenance}

Darwiche and Hirth (2020) connect sufficient reasons for classifier decisions to prime implicants, and later work extends beyond binary features (Ji and Darwiche, 2023). Receipts inherit minimal sufficiency and move it to typed, versioned source replacements in a multi-class transition. We specialize and extend their notion; minimal sufficiency itself is theirs.

The cube resembles actual-causality and provenance objects without being either. Halpern–Pearl causality asks whether events are causes under structural contingencies (Halpern and Pearl, 2005); provenance semirings track how outputs depend on input tuples (Green, Karvounarakis, and Tannen, 2007). A receipt is a minimal replay of a revised judgment from a specified old state, relative to named evaluator components; it asserts no metaphysical, legal, or moral cause.

\subsection{Model updates, selective prediction, and audit}

Model editing separates edit success from locality: an update should not disturb unrelated behavior (Mitchell et al., 2022; Ma et al., 2024). Drift and inertia give versioned evaluators the same decomposition, and receipts add attribution: a local, successful update can still be recorded under the wrong source. Explanation stability under retraining is related but different (Meyer et al., 2023): a receipt may properly change when a rule changes; the test is whether the change obeys the declared source semantics.

Selective prediction trades coverage against error under a reject option (Geifman and El-Yaniv, 2019). Reason debt is that discipline applied to accepted evaluator changes, paired with gold-change recall so low debt cannot be bought by refusing every amended case.

Paired transformations with known output relations are metamorphic tests (Chen et al., 1998; Segura et al., 2016), and the invariance cases parallel behavioral test suites for NLP models (Ribeiro et al., 2020). Our contribution is to attach these relations to an executable version-transition semantics, so a violated relation certifies a prediction error rather than flagging a heuristic concern.

Accountable-algorithm and internal-audit research asks for verifiable properties and lifecycle records, not source disclosure alone (Kroll et al., 2017; Raji et al., 2020). A judgment receipt is one narrow machine-checkable artifact inside that program.

\section{Scope and limitations}

\textbf{Constructed adjudicators.} ReasonBench has exact counterfactuals because its adjudicators are executable. That is the method's strength and its external-validity limit in one: real institutions carry tacit practice, conflicting records, discretionary exceptions, and contested authority.

\textbf{A declared source interface.} The accounting is only as complete as the interface. If policy hides in model weights, if authority changes what evidence means, or if retrieval and adjudication are entangled, three portable components can miss a real source. The abstraction must be audited, not assumed (Selbst et al., 2019).

\textbf{Hybrid coherence.} The cube assumes old and revised components share a typed contract and recombine without changing meaning. True in ReasonBench; not guaranteed across database migrations, ontology changes, or redesigned procedures. Such transitions need an explicit compatibility map or should be recorded as uncertifiable under the old interface.

\textbf{Determinism.} The formal results assume a deterministic adjudicator. Stochastic evaluators would need distributional receipts or a coupling that defines reproduction of an outcome; we do not solve that here.

\textbf{Exponential certification.} Exact unrestricted certification costs \(2^k\) cells. Three sources are practical; fine-grained decompositions are not. Monotonicity, sparsity, or verified program structure could cut query cost, but only as stated, tested assumptions.

\textbf{Benchmark composition.} The organizational locked test has six independent clauses; the logical stratum is large but templated. Aggregates can hide the imbalance, as the main cube deficit shows. We report stratum effects and do not treat six clauses as organizational policy at large.

\textbf{One model family.} The primary experiment uses one 1.7B open-weight backbone and one LoRA configuration; the 0.6B replication tests capacity in one direction only. The post-hoc study of \S\ref{sec:permaug} tests the permutation-augmentation route at the same scale, with three seeds and one augmentation rate. The direct-versus-cube comparison is also entangled with one serialization per target, a fixed cell order, and ordinary autoregressive decoding; keyed or per-cell cube supervision and constrained decoding are untested. Larger models and architectures that execute structured programs remain open comparisons.

\textbf{No human-validity study.} Receipts are exact under the benchmark semantics. Whether practitioners find the interface complete, whether auditors want the whole minimal family, and whether receipts improve appeals are untested questions.

\textbf{No two-plane systems trial.} The prediction/certification separation follows from the formal and predictive results; no deployed system was compared. Execution latency, audit scheduling, reviewer time, and saved cost are unmeasured.

\textbf{Authority is descriptive.} An authority flag records what the constructed evaluator may apply; it does not certify that the permission is legitimate. A receipt can make an appeal more precise; it cannot replace the appeal.

\section{Conclusion}

A versioned evaluator should not be judged by its revised answer alone. The transition has an account: accepted grounds, governing norms, and the authority that lets those norms speak. Executing their old/new hybrids yields a small counterfactual table from which every minimal account is recoverable.

The empirical lesson is less convenient than the frozen hypothesis. Predicting the full table does not beat predicting the receipt, and the result repeats at smaller scale. Worse, both targets look nearly perfect on a locked test while failing equivalent source orders and inverse transitions, and under unseen multi-source composition the verdicts survive while the receipts approach zero. A structured output is not structured understanding. Permutation augmentation largely restores the one relation it targets and leaves the next one broken; the structured target pays the largest price, in the stratum where it was already weakest.

The evidence supports a two-part standard. Certify reasons by executing counterfactuals; treat learned evaluator outputs as predictions whose consistency, coverage, and reason debt are measured against those certificates. ``No judgment without a reason'' is not a request for more fluent explanation. It requires a changed institutional action to stay replayable from an inspectable change in its declared grounds, rules, or authority.

\section*{References}

\addcontentsline{toc}{section}{References}

\footnotesize

Adebayo, Julius, Justin Gilmer, Michael Muelly, Ian Goodfellow, Moritz Hardt, and Been Kim. 2018. “Sanity Checks for Saliency Maps.” \emph{Advances in Neural Information Processing Systems 31}.

Bai, Yuntao, Saurav Kadavath, Sandipan Kundu, Amanda Askell, Jackson Kernion, et al. 2022. “Constitutional AI: Harmlessness from AI Feedback.” arXiv:2212.08073.

Barres, Victor, Honghua Dong, Soham Ray, Xujie Si, and Karthik Narasimhan. 2025. “\(\tau^2\)-Bench: Evaluating Conversational Agents in a Dual-Control Environment.” arXiv:2506.07982.

Bowman, Samuel R., Jeeyoon Hyun, Ethan Perez, Edwin Chen, Craig Pettit, et al. 2022. “Measuring Progress on Scalable Oversight for Large Language Models.” arXiv:2211.03540.

Chen, Tsong Yueh, Shing Chi Cheung, and Siu Ming Yiu. 1998. “Metamorphic Testing: A New Approach for Generating Next Test Cases.” Technical Report HKUST-CS98-01, Hong Kong University of Science and Technology.

Clark, Peter, Oyvind Tafjord, and Kyle Richardson. 2020. “Transformers as Soft Reasoners over Language.” In \emph{Proceedings of IJCAI 2020}, 3882–3890. \url{https://doi.org/10.24963/ijcai.2020/537}.

Darwiche, Adnan, and Auguste Hirth. 2020. “On the Reasons Behind Decisions.” In \emph{ECAI 2020}, 712–720. \url{https://doi.org/10.3233/FAIA200158}.

Dehghanighobadi, Zahra, Asja Fischer, and Muhammad Bilal Zafar. 2025. “Can LLMs Explain Themselves Counterfactually?” In \emph{Proceedings of EMNLP 2025}, 7787–7815. \url{https://doi.org/10.18653/v1/2025.emnlp-main.396}.

DeYoung, Jay, Sarthak Jain, Nazneen Fatema Rajani, Eric Lehman, Caiming Xiong, Richard Socher, and Byron C. Wallace. 2020. “ERASER: A Benchmark to Evaluate Rationalized NLP Models.” In \emph{Proceedings of ACL 2020}, 4443–4458. \url{https://doi.org/10.18653/v1/2020.acl-main.408}.

Engel, Konrad. 1997. \emph{Sperner Theory}. Cambridge University Press.

Geifman, Yonatan, and Ran El-Yaniv. 2019. “SelectiveNet: A Deep Neural Network with an Integrated Reject Option.” In \emph{Proceedings of ICML 2019}, 2151–2159. \url{https://proceedings.mlr.press/v97/geifman19a.html}.

Geirhos, Robert, Jörn-Henrik Jacobsen, Claudio Michaelis, Richard Zemel, Wieland Brendel, Matthias Bethge, and Felix A. Wichmann. 2020. “Shortcut Learning in Deep Neural Networks.” \emph{Nature Machine Intelligence} 2: 665–673. \url{https://doi.org/10.1038/s42256-020-00257-z}.

Green, Todd J., Grigoris Karvounarakis, and Val Tannen. 2007. “Provenance Semirings.” In \emph{Proceedings of PODS 2007}, 31–40. \url{https://doi.org/10.1145/1265530.1265535}.

Halpern, Joseph Y., and Judea Pearl. 2005. “Causes and Explanations: A Structural-Model Approach. Part I: Causes.” \emph{British Journal for the Philosophy of Science} 56(4): 843–887. \url{https://doi.org/10.1093/bjps/axi147}.

Hu, Edward J., Yelong Shen, Phillip Wallis, Zeyuan Allen-Zhu, Yuanzhi Li, Shean Wang, Lu Wang, and Weizhu Chen. 2022. “LoRA: Low-Rank Adaptation of Large Language Models.” \emph{International Conference on Learning Representations}. \url{https://openreview.net/forum?id=nZeVKeeFYf9}.

Irving, Geoffrey, Paul Christiano, and Dario Amodei. 2018. “AI Safety via Debate.” arXiv:1805.00899.

Jacovi, Alon, and Yoav Goldberg. 2020. “Towards Faithfully Interpretable NLP Systems: How Should We Define and Evaluate Faithfulness?” In \emph{Proceedings of ACL 2020}, 4198–4205. \url{https://doi.org/10.18653/v1/2020.acl-main.386}.

Ji, Chunxi, and Adnan Darwiche. 2023. “A New Class of Explanations for Classifiers with Non-Binary Features.” In \emph{Logics in Artificial Intelligence (JELIA 2023)}, Lecture Notes in Computer Science 14281, 106–122. \url{https://doi.org/10.1007/978-3-031-43619-2_8}.

Kroll, Joshua A., Joanna Huey, Solon Barocas, Edward W. Felten, Joel R. Reidenberg, David G. Robinson, and Harlan Yu. 2017. “Accountable Algorithms.” \emph{University of Pennsylvania Law Review} 165(3): 633–705. \url{https://pennlawreview.com/2017/02/23/accountable-algorithms/}.

Lanham, Tamera, Anna Chen, Ansh Radhakrishnan, Benoit Steiner, Carson Denison, et al. 2023. “Measuring Faithfulness in Chain-of-Thought Reasoning.” arXiv:2307.13702.

Lightman, Hunter, Vineet Kosaraju, Yura Burda, Harri Edwards, Bowen Baker, Teddy Lee, Jan Leike, John Schulman, Ilya Sutskever, and Karl Cobbe. 2024. “Let's Verify Step by Step.” \emph{International Conference on Learning Representations}. arXiv:2305.20050.

Liu, Nelson F., Kevin Lin, John Hewitt, Ashwin Paranjape, Michele Bevilacqua, Fabio Petroni, and Percy Liang. 2024. “Lost in the Middle: How Language Models Use Long Contexts.” \emph{Transactions of the Association for Computational Linguistics} 12: 157–173. \url{https://doi.org/10.1162/tacl_a_00638}.

Ma, Jun-Yu, Zhen-Hua Ling, Ningyu Zhang, and Jia-Chen Gu. 2024. “Neighboring Perturbations of Knowledge Editing on Large Language Models.” In \emph{Proceedings of ICML 2024}, 33839–33854. \url{https://proceedings.mlr.press/v235/ma24h.html}.

Mayne, Harry, Ryan Othniel Kearns, Yushi Yang, Andrew M. Bean, Eoin D. Delaney, Chris Russell, and Adam Mahdi. 2025. “LLMs Don't Know Their Own Decision Boundaries: The Unreliability of Self-Generated Counterfactual Explanations.” In \emph{Proceedings of EMNLP 2025}, 24161–24186. \url{https://doi.org/10.18653/v1/2025.emnlp-main.1231}.

Meyer, Anna P., Dan Ley, Suraj Srinivas, and Himabindu Lakkaraju. 2023. “On Minimizing the Impact of Dataset Shifts on Actionable Explanations.” In \emph{Proceedings of UAI 2023}, 1434–1444. \url{https://proceedings.mlr.press/v216/meyer23a.html}.

Miller, Tim. 2019. “Explanation in Artificial Intelligence: Insights from the Social Sciences.” \emph{Artificial Intelligence} 267: 1–38. \url{https://doi.org/10.1016/j.artint.2018.07.007}.

Mitchell, Eric, Charles Lin, Antoine Bosselut, Christopher D. Manning, and Chelsea Finn. 2022. “Memory-Based Model Editing at Scale.” In \emph{Proceedings of ICML 2022}, 15817–15831. \url{https://proceedings.mlr.press/v162/mitchell22a.html}.

Ouyang, Long, Jeffrey Wu, Xu Jiang, Diogo Almeida, Carroll Wainwright, et al. 2022. “Training Language Models to Follow Instructions with Human Feedback.” \emph{Advances in Neural Information Processing Systems 35}, 27730–27744.

Raji, Inioluwa Deborah, Andrew Smart, Rebecca N. White, Margaret Mitchell, Timnit Gebru, Ben Hutchinson, Jamila Smith-Loud, Daniel Theron, and Parker Barnes. 2020. “Closing the AI Accountability Gap: Defining an End-to-End Framework for Internal Algorithmic Auditing.” In \emph{Proceedings of the 2020 Conference on Fairness, Accountability, and Transparency}, 33–44. \url{https://doi.org/10.1145/3351095.3372873}.

Ribeiro, Marco Tulio, Tongshuang Wu, Carlos Guestrin, and Sameer Singh. 2020. “Beyond Accuracy: Behavioral Testing of NLP Models with CheckList.” In \emph{Proceedings of ACL 2020}, 4902–4912. \url{https://doi.org/10.18653/v1/2020.acl-main.442}.

Ross, Andrew Slavin, Michael C. Hughes, and Finale Doshi-Velez. 2017. “Right for the Right Reasons: Training Differentiable Models by Constraining their Explanations.” In \emph{Proceedings of IJCAI 2017}, 2662–2670. \url{https://doi.org/10.24963/ijcai.2017/371}.

Segura, Sergio, Gordon Fraser, Ana B. Sanchez, and Antonio Ruiz-Cortés. 2016. “A Survey on Metamorphic Testing.” \emph{IEEE Transactions on Software Engineering} 42(9): 805–824. \url{https://doi.org/10.1109/TSE.2016.2532875}.

Selbst, Andrew D., Danah Boyd, Sorelle A. Friedler, Suresh Venkatasubramanian, and Janet Vertesi. 2019. “Fairness and Abstraction in Sociotechnical Systems.” In \emph{Proceedings of the Conference on Fairness, Accountability, and Transparency}, 59–68. \url{https://doi.org/10.1145/3287560.3287598}.

Shi, Haochen, Shaobo Li, Guoqing Chao, Xiaoliang Shi, Wentao Chen, and Zhenzhou Ji. 2025. “Do LLMs Behave as Claimed? Investigating How LLMs Follow Their Own Claims Using Counterfactual Questions.” In \emph{Proceedings of EMNLP 2025}, 29043–29056. \url{https://doi.org/10.18653/v1/2025.emnlp-main.1479}.

Turpin, Miles, Julian Michael, Ethan Perez, and Samuel R. Bowman. 2023. “Language Models Don't Always Say What They Think: Unfaithful Explanations in Chain-of-Thought Prompting.” \emph{Advances in Neural Information Processing Systems 36}.

Yang, An, et al. 2025. “Qwen3 Technical Report.” arXiv:2505.09388. \url{https://arxiv.org/abs/2505.09388}.

Zhang, Weining. 2026. “Eval Is an Institution: Witness-Gated Patching for Versioned AI Judgment.” Companion manuscript.

Zheng, Lianmin, Wei-Lin Chiang, Ying Sheng, Siyuan Zhuang, Zhanghao Wu, et al. 2023. “Judging LLM-as-a-Judge with MT-Bench and Chatbot Arena.” \emph{Advances in Neural Information Processing Systems 36}, Datasets and Benchmarks Track.

\end{document}